%% file: bmvc_final.tex
\documentclass{bmvc2k}

\title{A Layer-wise Adversarial-aware Quantization Optimization for Improving Robustness}

\addauthor{Chang Song}{chang.song@duke.edu}{1}
\addauthor{Riya Ranjan}{riya.ranjan.00@gmail.com}{2}
\addauthor{Hai Li}{hai.li@duke.edu}{1}

\addinstitution{
 Department of ECE\\
 Duke University\\
 Durham, NC, USA
}
\addinstitution{
 Monta Vista High School\\
 21840 McClellan Rd,\\
 Cupertino, CA, USA
}

\runninghead{Song, Ranjan, Li}{Layer-wise Adversarial-aware Quantization}

\def\etal{\emph{et al}\bmvaOneDot}

\usepackage{booktabs}
\usepackage[ruled,vlined]{algorithm2e}
\usepackage{adjustbox}
\usepackage{multirow}
\usepackage{enumitem}

\usepackage{color}
\pagecolor{white}

\begin{document}
\emergencystretch 3em

\maketitle

\vspace{-6mm}
\begin{abstract}
Neural networks are getting better accuracy with higher energy and computational cost. After quantization, the cost can be greatly saved, and the quantized models are more hardware friendly with acceptable accuracy loss. On the other hand, recent research has found that neural networks are vulnerable to adversarial attacks, and the robustness of a neural network model can only be improved with defense methods, such as adversarial training. In this work, we find that adversarially-trained neural networks are more vulnerable to quantization loss than plain models. To minimize both the adversarial and the quantization losses simultaneously and to make the quantized model robust, we propose a layer-wise adversarial-aware quantization method, using the Lipschitz constant to choose the best quantization parameter settings for a neural network. We theoretically derive the losses and prove the consistency of our metric selection. The experiment results show that our method can effectively and efficiently improve the robustness of quantized adversarially-trained neural networks.
\end{abstract}
\vspace{-3mm}

\input{01_intro}
\input{02_background}
\input{03_motivation}
\input{04_theory}
\input{05_method}
\input{06_experiment}
\input{07_conclusions}

\bibliography{egbib}
\end{document}

%% file: 01_intro.tex
\section{Introduction}
\label{intro}

Neural networks have achieved rapid development with a variety of applications~\cite{lecun2015deep}. With more complex structures, neural network models demonstrate near or even beyond human-level accuracy in classification problems~\cite{graham2014fractional}. Neural networks have not only become the main trend in image recognition, object detection, speech recognition, and natural language processing tasks, but also have been widely adopted by the security industry~\cite{tao2006human}, including use in surveillance, authentication, facial recognition, and crowd behavior analysis.

With the assistance of rapid development of hardware platforms, neural network architecture has evolved dramatically, handling more complicated datasets and achieving better performance. 
From initially shallow structures like LeNet~\cite{lecun1998gradient} and AlexNet~\cite{krizhevsky2012imagenet}, to deeper and wider designs such as GoogLeNet (a.k.a Inception)~\cite{szegedy2015going}, VGG~\cite{simonyan2014very} and ResNet~\cite{he2016deep}, the parameters of neural network models have grown exponentially from 60 thousand to over 100 million, and the total multiply-accumulate operations (MACs) has increased from 341 thousand to 15 billion. 
Such explosions in model sizes and MACs require large computation resources and memory storage and result in high power consumption.
This makes the deployment of neural networks very challenging, especially on low-power mobile devices, such as autonomous vehicle systems~\cite{maqueda2018event} and real-time robot navigation systems~\cite{ribeiro2017real}. 
In order to fit neural networks into such resource-constrained hardware platforms, model compression methods such as quantization~\cite{hubara2017quantized}, sparsification~\cite{wen2016learning,liu2015sparse}, and pruning~\cite{han2015learning} have been extensively studied to save memory and computation costs and improve execution speed while maintaining performance. For example, high cost high performance neural networks such as DNNs have been optimized using quantization techniques that both improve accuracy and cost efficiency~\cite{chen2021quantization}. These techniques are highly valuable, as they can then be applied to high level processing that is used in a variety of real world applications.

In addition to execution efficiency, the deep learning research community pays close attention to security and privacy problems, particularly. 
Recent research shows that neural network models are vulnerable to certain attacks, especially the so-called adversarial attacks~\cite{szegedy2013intriguing}. 
Adversarial attacks can easily fool models and lead to misclassifications by elaborately generating imperceptible perturbations added to inputs (a.k.a. adversarial examples). 
% Moreover, adversarial attacks are transferable, that is, adversarial examples generated by a model can even influence other models with different structures~\cite{papernot2016transferability}. 
% Such imperceptible attacks with great transferability unveil the blind spots in neural networks and raise severe reliability and security concerns.
As the research on machine learning and neural networks goes deeper, different types of adversarial attacks are proposed.
However, most defenses are effective at mitigating only one specific attack, rendering them ineffective as more attacks are created. 
For example, Athalye \emph{et al.} claim that the attacks based on the obfuscated gradients they proposed could circumvent 7 out of 9 noncertified white-box-secure defenses that were accepted by the
% \textit{International Conference on Learning Representations (ICLR)} 2018
ICLR 2018~\cite{athalye2018obfuscated}. 
To better address the universality issue and to fully analyze adversarial attacks, some researchers focus on model robustness rather than solely relying on accuracy of certain attacks. Many researchers adversarially train neural networks to increase their models' robustness~\cite{athalye2018obfuscated}. However, the use of this defense mechanism also increases the loss when quantizing such adversarially trained networks. Therefore, a need exists for a quantization method that does not compromise the accuracy of adversarially trained neural networks.

In this work, we aim to improve the robustness of weight-quantized neural networks. 
Compared to existing works on defending adversarial attacks and improving neural network robustness, our major contributions include: (1) We find that with careful quantization setting selection, the robustness of quantized adversarially-trained models can be recovered back to the same level as their full-precision counterparts. (2) We propose a layer-wise adversarial-aware quantization method to minimize the robustness loss introduced by adversarial attacks and quantization. (3) We theoretically prove the effectiveness of the layer-wise adversarial-aware quantization method and derive the metrics for error sensitivity. (4) Experimental results show that the robustness of quantized adversarially-trained models can be restored by using the layer-wise adversarial-aware quantization method.
% \vspace{-2mm}
% \begin{itemize}[leftmargin=*]
%     \item We find that with careful quantization setting selection, the robustness of quantized adversarially-trained models can be recovered back to the same level as their full-precision counterparts.
%     \vspace{-2mm}
% 	\item We propose a layer-wise adversarial-aware quantization method to minimize the robustness loss introduced by adversarial attacks and quantization.
% 	\vspace{-2mm}
% 	\item We theoretically prove the effectiveness of the layer-wise adversarial-aware quantization method and derive the metrics for error sensitivity.
% 	\vspace{-2mm}
% 	\item We test our method and the results show that the robustness of quantized adversarially-trained models can be restored by using the layer-wise adversarial-aware quantization method.
% 	\vspace{-2mm}
% \end{itemize}

%% file: 02_background.tex
% \vspace{-4mm}
\section{Background}
\label{background}

% \vspace{-2mm}
\subsection{Adversarial Attacks and Robustness}

Adversarial attacks generate adversarial examples to fool models. 
An adversarial example $\widetilde{X}$ is generated by injecting adversarial perturbation $\epsilon$ (a.k.a. adversarial strength) to a clean sample $X$: $\widetilde{X}=X+\epsilon$.
% \begin{equation}
%     \label{eq:0}
%     % \small
%     \widetilde{X}=X+\epsilon.
% \end{equation}
Usually, adversarial perturbations are so tiny that they are even imperceptible to human eyes. 
However, carefully designed adversarial perturbations can cause a neural network to misclassify adversarial examples with high confidence levels~\cite{goodfellow2014explaining}. 
There are multiple methods to generate $\epsilon$, such as FGSM~\cite{goodfellow2014explaining}, CW~\cite{carlini2017towards}, and PGD~\cite{madry2017towards}.
% $\epsilon$ is also referred to as adversarial strength; larger $\epsilon$ means $\widetilde{X}$ is further away from $X$ in the decision space. 
% Adversarial attacks raise security concerns about the applications of neural networks in real-world scenarios. 
% For example, attackers could alter a ``stop'' sign imperceptibly and make an autonomous vehicle interpret as a ``yield'' sign~\cite{papernot2017practical}.

According to the attackers' knowledge, adversarial attacks can be classified as \textit{black-box}, \textit{gray-box}, or \textit{white-box} attacks. 
The attackers in black-box settings can access only the inputs and outputs of the target model and they do not know the target model's structure and parameters. 
% Thus in black box scenarios the attackers usually design and train their own model to generate adversarial examples. 
In gray-box attacks, the attackers only know the target model's structure, while in white-box attacks they have a full access to the target model.
% , and therefore can create specific adversarial examples against the target model.
%
A given model usually has worse performance against white-box attacks than gray- and black-box attacks (with the same attack strengths), as the attacker has more knowledge about the model in white-box scenarios~\cite{papernot2016transferability}. 
% Due to the model transferability~\cite{papernot2016transferability}, black-box attacks are feasible but they usually need larger adversarial strengths to downgrade accuracy compared to white-box attacks. 
% {\RED But relatively speaking, white-box attacks are easier to defend than black-box attacks. [I still concern this. Why??]}
Other key adversarial definitions include \textit{targeted}/\textit{non-targeted} attacks, \textit{iterative}/\textit{non-iterative} attacks~\cite{yuan2019adversarial}. 
% The goal of targeted attacks is to fool a model to classify examples in one class as a specific class, while non-targeted attacks aim to increase the model's misclassification rate but do not care the classification results. 
% Iterative attacks perform multiple steps of attacks to optimize the final adversarial examples and non-iterative attacks only perturb the samples once.
In this paper, we mainly focus on non-targeted attacks and we will give equal consideration to gray- and white-box, iterative and non-iterative attacks.

% \vspace{-3mm}
% \subsection{Robustness}

Robustness indicates how well models can tolerate different types of perturbations and retain proper function. 
Before adversarial attacks were discovered, robustness was discussed mainly in random noise scenarios~\cite{ulivcny2016robustness}. 
Adversarial perturbations $\epsilon$ can be regarded as the worst-case noise which result in large classification errors with minimum effort. 
In this work, we consider the overall test accuracy of multiple types of adversarial examples as a robustness indicator: \textit{If a model can achieve an overall high accuracy against multiple types of attacks~\cite{athalye2018obfuscated}, the model is robust.}
So far, \textit{adversarial training} is taken as the most effective and efficient way to mitigate the damage of adversarial attacks and to improve model robustness.
This training method includes adversarial examples in the training set to enhance the models’ resilience to adversarial attacks. 
The main idea of adversarial training is similar to data augmentation: to facilitate training by providing more information on data distribution. 
The effectiveness of adversarial training has been demonstrated and explained by~\cite{goodfellow2014explaining}. 
% Note that in adversarial training, the model that is used to generate the adversarial examples is not necessarily identical to the model being attacked. 
Some other popular methods that have been studied to improve the robustness of neural networks are \textit{gradient masking}~\cite{papernot2017practical} and \textit{defensive distillation}~\cite{soll2019evaluating}.
% Some other popular methods have been studied to improve the robustness of neural networks. 
% Generally, these methods can be cataloged as two types: (1) Gradient masking: which essentially builds a model to hide or smooth the gradient between original and adversarial examples~\cite{papernot2017practical}.
% It is particularly effective against white-box attacks; however, when the attacker uses a model different from the protected model, gradient masking is ineffective~\cite{athalye2018obfuscated}.
% (2) Defensive distillation: which creates a model whose decision boundaries are smoothed along the directions that the attacker may exploit. Defensive distillation makes it difficult for the attacker to discover adversarial input tweaks that lead to incorrect classes~\cite{soll2019evaluating}.

% \vspace{-4mm}
\subsection{Quantization}

The recent performance enhancement of neural networks mainly benefits from more parameters and operations. Such costs are not tolerable for limited-power or real-time applications. 
Neural network quantization~\cite{fiesler1990weight} helps save computation and memory costs and therefore improves power efficiency. Quantization has also been shown to improve accuracy in some cases, and in almost all cases retains the accuracy of the original network.
Moreover, quantization enables the deployment of neural networks on limited-precision hardware. 
Many hardware systems have physical constraints and do not allow full-precision models to directly map to them. 
For example, GPUs support half precision floating point arithmetic (FP16); ReRAM (a.k.a memristor)~\cite{strukov2008missing} only allows limited precision because of process variations~\cite{6709674,10.1145/2744769.2744900}.
% There are two types of quantization methods: \textit{post-training quantization} and \textit{quantization-aware training}. 
% Post-training quantization directly quantizes a model without retraining. However, recent research show that 8-bit post-training quantization does not preserve accuracy in some cases, especially for small models~\cite{jacob2018quantization}. 
% Quantization-aware training emulates inference-time quantization, and keeps a copy of full-precision weights to accumulate gradient changes. 
% During the backpropagation, the quantization functions are discrete-valued and cannot produce derivatives for training. 
% A popular bypass is using the straight-through estimator (STE) to approximate the backpropagation process~\cite{bengio2013estimating}.
Quantization levels can be linear or logarithmic~\cite{miyashita2016convolutional}, 
% In linear quantization, quantization levels are uniformly distributed. While in logarithmic quantization, quantization levels follow logarithmic functions. 
% Both of them are feasible and show similar effect.
and quantization can be applied to weight parameters as well as activations. 
%There are two types of quantization as for the parts of neural networks being quantized: weight-quantization and activation-quantization. 
It has been observed that activations are more sensitive to quantization than weights~\cite{zhou2016dorefa}. 
Some works quantize activations to a higher precision than weights, some only quantize weights and keep activation at full-precision, and only a few studies explore both~\cite{8999057}. 
We consider the weight-quantization only as we agree that quantization does more harm to activations than to weights. 

Recent works related with robustness and quantization mainly focus on two different directions: using quantization as a defense method or improving adversarial robustness in quantization scenarios. For defending adversarial attacks using quantization, one recent work~\cite{galloway2017attacking} shows that quantization can defend adversarial attacks through gradient masking, though the effectiveness is limited. In another work~\cite{rakin2018defend}, the authors use quantized activation functions to defend adversarial attacks. 
For improving adversarial robustness in quantized neural networks, in one of the related works called QUANOS~\cite{panda2020quanos}, the authors perform layer-specific hybrid quantization on a neural network while optimally trading off between energy-accuracy-and-robustness. To be specific, the authors use a metric called Adversarial Noise Sensitivity (ANS) to measure each layer's sensitivity to adversarial noise, and then quantize each layer in proportion to its ANS value. However, the authors only use white-box FGSM attack to calculate the ANS metric, and use white-box FGSM and PGD attacks to verify the robustness of QUANOS, which are not enough. What is more, the authors fail to consider the quantization loss of QUANOS. Therefore there exists a need for a method of determining optimal quantization parameters that is effective against both white-box and gray-box attacks, and minimizes both quantization and adversarial losses.

%% file: 03_motivation.tex
% \vspace{-5mm}
\section{Motivation}
\label{motivation}

% \vspace{-2mm}
\subsection{Adversarial and Quantization Losses}

For a single fully-connected layer, suppose $W$ is the weight matrix, $W+\Delta W$ is the weight after quantization, $x$ is the original input, and $x+\Delta x$ is the adversarial input. The difference in the output of this layer ($\delta$) can be represented as follow:
% \vspace{-2mm}
\begin{equation}
    \label{eq:1_Q}
    % \small
    \delta = (W+\Delta W) \cdot (x+\Delta x)-Wx = %\underbrace{W\Delta x}_{\text{Adv.}} + \underbrace{\Delta Wx}_{\text{Quant.}} + \Delta W\Delta x.
    W\Delta x + \Delta Wx + \Delta W\Delta x.
    % \vspace{-2mm}
\end{equation}
The first term on the right hand side of Eq.~(\ref{eq:1_Q}) is the adversarial loss introduced by adversarial examples, the second term is the quantization loss introduced by weight quantization, and the third term is the loss introduced by adversarial examples and quantization together. Since the third term is much smaller than the first two, we can omit it for simplicity.

The adversarial loss can be easily measured by the accuracy drop of a model under adversarial attacks. For quantization loss, the accuracy drop could reflect the change in weights, but not directly. 
The difference in outputs between a full-precision model and its quantized counterpart reflect the effect of quantization, but it also depends on the input. 
Thus, an input-independent criterion is essentially necessary to evaluate the quantization process.

% \vspace{-2mm}
\subsection{The Error Amplification Effect and the Lipschitz Constant}

The error amplification effect was initially introduced in Liao \etal~\cite{liao2018defense}. 
It indicates the situation that small residual perturbation is amplified to a large magnitude in top layers of a model and finally leads to a wrong prediction. 
This effect mainly focuses on the input-level perturbation, especially adversarial perturbations. 
In this work, we focus more on the weight-level error. 
Similar to its original definition, \textbf{the quantization error amplification effect} means that each layer will inject a quantization loss. 
The error introduced by quantization can be amplified layer by layer and eventaully results in misclassification in the output.

To measure the influence of quantization loss, we propose to use \textbf{the Lipschitz constant} as a quantization error amplification indicator. 
Based on Cisse \etal~\cite{cisse2017parseval}, the Lipschitz constant of the weight loss ($\Delta W$) introduced by quantization can be defined as:
\vspace{-2mm}
\begin{equation}
    \label{eq:2_Q}
    % \small
    \left \| \Delta W \right \|_p = \sup_{z: \left \| z \right \|_p = 1} \left \| \Delta Wz \right \|_p ,
    \vspace{-2mm}
\end{equation}

where, $z$ is an arbitrary normalized input. 
Eq.~(\ref{eq:2_Q}) shows that the quantization loss can be upper-bounded by the Lipschitz constant of $\Delta W$. 
A Lipschitz constant greater than 1 indicates that the weight undergoes a large quantization loss; a Lipschitz constant less than 1 means the weight has a small loss after quantization.
$\left \| \Delta W \right \|_2$ when $p=2$ is called the spectral norm of $\Delta W$, which is the maximum singular value of $\Delta W$. 
$\left \| \Delta W \right \|_\infty$ when $p=+\infty$ is the maximum 1-norm of the rows of $\Delta W$. The Lipschitz constant additionally provides an input-independent criterion that will allow us to evaluate quantization loss and determine the effectiveness of our quantization process, given that it is based on weights
rather than inputs.

All the layers aforementioned are fully-connected layers, which perform basic matrix-vector or matrix-matrix multiplications. For more complicated layer structures, such as convolutional layers, Cisse \etal \cite{cisse2017parseval} has already given detailed explanations on how to represent convolution operations as basic matrix multiplications and we have verified the correctness of their derivations. 

\vspace{-6mm}
\subsection{Adversarial Training and Quantization}
% \vspace{-2mm}

\begin{figure*}[tb]
    % \vspace{-1mm}
	\centering
	\includegraphics[width=.9\linewidth]{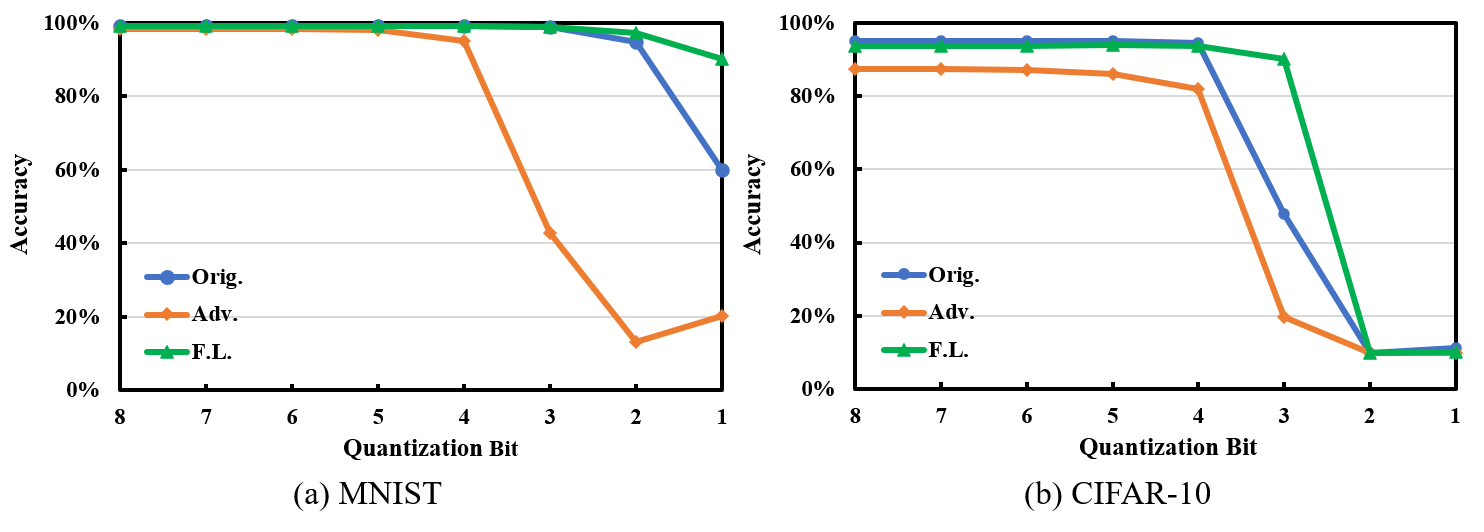}
	\vspace{-4mm}
	\caption{Clean Accuracy vs. Quantization Levels. Orig.: a vanilla model, Adv.: an adversarially-trained model, F.L.: a model trained with feedback learning~\cite{8999302}.}
	\label{fig:motivation}
	\vspace{-4mm}
\end{figure*}

As discovered in Lin \etal~\cite{lin2019defensive}, quantized models are more vulnerable to adversarial attacks than their original full-precision models. 
We reproduced their experiments and obtained similar results. 
Furthermore, we find that adversarially-trained models experience greater loss from quantization than other models. As shown in Figure~\ref{fig:motivation} (refer to Section~\ref{experiment} for the setup details), the accuracy gap between full-precision adversarially-trained models and quantized counterparts is larger than other models. Here, Orig. is a vanilla model with standard training methods, Adv. is adversarially-trained by Madry \etal~\cite{madry2017towards}, F.L. is retrained from the original model using feedback learning, as proposed in Song \etal~\cite{8999302}. Although methods like feedback learning can minimize the quantization loss, the most effective and efficient way to improve adversarial robustness is still adversarial training. So quantizing adversarially-trained models with careful parameter selection is necessary to minimize both adversarial and quantization loss.

Figure~\ref{fig:motivation} shows that as the weight quantization level decreases, the accuracy of adversarially-trained models drops more quickly than that of other models. 
For both MNIST and CIFAR-10 datasets, the F.L. model has the best resistance to the weight quantization. 
The adversarial model is vulnerable to quantization loss, which has an even lower accuracy than the original model, especially when the precision level is low. 
For the MNIST dataset, the accuracy gap between the adversarial model and the other two models enlarges as the quantization level drops, which indicates again that adversarial training is highly vulnerable to quantization loss. 
In contrast, for the CIFAR-10 dataset, all three models depreciate at 1-bit and 2-bit quantization. 
The different performance of two datasets implies that complicated datasets and models need higher precision than simpler datasets and models. 
% In this work, we mainly focus on 3-bit quantization to get better a understanding of different training techniques.

% Here is a possible explanation of why adversarial training does not perform well in the quantized model. 
% Adversarial training introduces additional examples to the training set, which helps reshape decision boundaries to better defend adversarial attacks. However, it fails to compensate for the quantization effect on decision boundaries and even leads to larger quantization loss. 
% We will validate this explanation in \textit{Experimental Evaluations} using Lipschitz constant.

The robustness evaluation results on MNIST and CIFAR-10 datasets are given in Table~\ref{tab:MNIST_gray_both} and Table~\ref{tab:CIFAR_gray_both}, respectively. 
Here in both tables, ``Q'' means uniformly using 3-bit quantization for all layers, ``LQ'' means using a layer-wise quantization which gives selected layers additional quantization bits. We find that with careful quantization selection, the robustness of quantized adversarially-trained model can be greatly regained.

\begin{table}[tb]
% \vspace{1mm}
\centering
\caption{The robustness on MNIST models under gray-box attacks.}
\label{tab:MNIST_gray_both}
\resizebox{0.9\linewidth}{!}{%
\begin{tabular}{lrrrrrrr}
\toprule
Models      & Clean            & CW-L2            & FGSM (w)         & FGSM (s)         & PGD              & BIM              & MIM              \\ \cmidrule(lr){1-1} \cmidrule(lr){2-8}
Adv.          & 98.40\%          & 96.74\%          & 98.02\%          & 96.28\%          & 97.97\%          & 97.76\%          & 97.76\%          \\
Adv. (Q)      & 42.69\%          & 38.70\%          & 43.54\%          & 39.91\%          & 45.05\%          & 43.94\%          & 43.61\%          \\ \midrule
Adv. (LQ)     & 95.64\%          & 90.50\%          & 94.67\%          & 87.80\%           & 94.62\%           & 94.09\%           & 93.82\%           \\ \bottomrule
\end{tabular}%
}
% \vspace{-6mm}
\end{table}

\begin{table}[tb]
\vspace{-2mm}
\centering
\caption{The robustness of CIFAR-10 models under gray-box attacks.}
\label{tab:CIFAR_gray_both}
\resizebox{0.9\linewidth}{!}{%
\begin{tabular}{lrrrrrrr}
\toprule
Models      & Clean            & CW-L2            & FGSM (w)         & FGSM (s)         & PGD              & BIM              & MIM              \\ \cmidrule(lr){1-1} \cmidrule(lr){2-8}
Adv.          & 87.27\%          & 86.30\%          & 86.20\%          & 68.60\%          & 86.10\%          & 85.50\%          & 86.10\%          \\
Adv. (Q)      & 19.84\%          & 20.00\%          & 19.40\%          & 16.20\%          & 19.20\%          & 19.20\%          & 19.10\%          \\ \midrule
Adv. (LQ)     & 83.66\%          & 83.50\%          & 82.90\%          & 68.30\%           & 83.30\%           & 82.80\%           & 82.80\%           \\ \bottomrule
\end{tabular}%
}
\vspace{-4mm}
\end{table}

%% file: 04_theory.tex
% \vspace{-5mm}
\section{Theoretical Derivation and Proof}
\label{theoretical_both}

% \vspace{-2mm}
\subsection{Adversarial and Quantization Loss Derivation}

Recall the loss derivation in Eq.~(\ref{eq:1_Q}): the total loss is the sum of adversarial loss, quantization loss, and $\Delta W \Delta x$. Here we will further derive the loss in consideration of activation functions to validate the metrics we will use for our layer-wise quantization optimization.

Let us assume that for a neural network which has a total of $L$ layers, the input vector and the input  perturbation introduced by the adversarial noise of layer $l$ are $x^l$ and $\Delta x^l$, respectively (both are column vectors with $n$ elements). The full-precision weight matrix and the quantized weight matrix of layer $l$ are $W^l$ and $W^l+\Delta W^l$, respectively (both are $m \times n$ matrices). Here $\Delta W^l$ is the error introduced by quantization in layer $l$. Then the output of layer $l$ is:
\begin{equation}
    \label{eq:1_Both}
    % \small
    x_{a,q}^{l+1}=a^l[(W^l+\Delta W^l)(x^l+\Delta x^l)], 
    % \vspace{-1mm}
\end{equation}

where $a^l(\cdot)$ is the element-wise activation function of layer $l$: The $i$-th element of $x_{a,q}^{l+1}$ is $a^l[\sum_{j=1}^{n}(W^l_{ij}+\Delta W^l_{ij})(x^l_j+\Delta x^l_j)]$.
The loss introduced by adversarial and quantization loss in the output of layer $l$ is:
% \vspace{-2mm}
\begin{equation}
    \label{eq:2_Both}
    % \small
    \Delta x_{a,q}^{l+1}=x_{a,q}^{l+1}-x^{l+1}=a^l[(W^l+\Delta W^l)(x^l+\Delta x^l)]-a^l(W^l x^l). 
    % \vspace{-1mm}
\end{equation}

% According to Lagrange’s Mean Value Theorem, $\exists \xi \in (W^l x^l, (W^l+\Delta W^l)(x^l+\Delta x^l)), s.t.$:
% \vspace{-2mm}
% \begin{equation}
%     \label{eq:3_Both}
%     \Delta x_{a,q}^{l+1}=[a^l(\xi)]'(W^l \Delta x^l+\Delta W^l x^l+\Delta W^l \Delta x^l).
%     \vspace{-1mm}
% \end{equation}

The $i$-th element of $\Delta x_{a,q}^{l+1}$ is $a^l[\sum_{j=1}^n (W^l_{ij}+\Delta W^l_{ij})(x^l_j+\Delta x^l_j)]-a^l(\sum_{j=1}^n W^l_{ij}x^l_j)$. To get rid of the activation function, let us consider the overall error before activation:

% \vspace{-2mm}
\begin{equation}
    \label{eq:3_Both}
    (W^l+\Delta W^l)(x^l+\Delta x^l) - W^l x^l = W^l \Delta x^l+\Delta W^l x^l+\Delta W^l \Delta x^l.
    % \vspace{-1mm}
\end{equation}

Its $i$-th element can be written as $\sum_{j=1}^n (W^l_{ij} \Delta x^l_j+\Delta W^l_{ij} x^l_j+\Delta W^l_{ij} \Delta x^l_j).$

Commonly-used activation functions, such as ReLU, sigmoid, tanh, etc., are monotonically non-decreasing functions. Thus, the outputs of activation functions are proportional to their input values:

\vspace{-5mm}
\begin{equation}
    \label{eq:3.5_Both}
    a^l[\sum_{j=1}^n (W^l_{ij}+\Delta W^l_{ij})(x^l_j+\Delta x^l_j)]-a^l(\sum_{j=1}^n W^l_{ij}x^l_j) \propto \sum_{j=1}^n (W^l_{ij} \Delta x^l_j+\Delta W^l_{ij} x^l_j+\Delta W^l_{ij} \Delta x^l_j).
    \vspace{-2mm}
\end{equation}

% with limited range derivatives: $[a^l(\cdot)]' \in [0, 1]$. Thus:
Since each element of $\Delta x_{a,q}^{l+1}$ is proportional to the corresponding element of $(W^l \Delta x^l+\Delta W^l x^l+\Delta W^l \Delta x^l)$, we have:

\vspace{-2mm}
\begin{equation}
    \label{eq:4_Both}
    \Delta x_{a,q}^{l+1} \propto W^l \Delta x^l+\Delta W^l x^l+\Delta W^l \Delta x^l.
    % \vspace{-2mm}
\end{equation}

Here, we define the adversarial loss $\Delta x_{a}^{l+1}=W^l \Delta x^l$, the quantization loss $\Delta x_{q}^{l+1}=\Delta W^l x^l$, and the mutual loss is $\Delta W^l \Delta x^l$.
For \textbf{the adversarial loss $\Delta x_{a}^{l+1}$}, if we treat $\Delta x^l$ as the input error and $\Delta x_{a}^{l+1}$ as the output error, $W^l$ will be the amplification factor that enlarges the error. An effective and efficient way to measure the adversarial loss amplification effect is the Lipschitz constant of $W^l$.
For \textbf{the quantization loss $\Delta x_{q}^{l+1}$}, we can not guarantee the input $x^l$ is bounded, but if we consider it as a constant value matrix or vector, then bounding the quantization loss is equivalent to bounding $\Delta W^l$.
For \textbf{the mutual loss}, we will use the induced p-norm to bound the error. As all induced norms are consistent, with $\left \| \Delta x^l \right \|_{p} \leq \epsilon^l$ (when $l=1$, $\epsilon^1$ is the adversarial strength of an adversarial example):
\vspace{-2mm}
\begin{equation}
    \label{eq:5_Both}
    \left \| \Delta W^l \Delta x^l \right \|_{p} \leq \left \| \Delta W^l \right \|_{p} \left \| \Delta x^l \right \|_{p} \leq \left \| \Delta W^l \right \|_{p} \epsilon^l.
    \vspace{-2mm}
\end{equation}

As we have already bounded $\Delta W^l$ for the quantization loss, $\epsilon^l$ is also bounded as $\epsilon^1$ has negligible strength compared to $\left \| x^1 \right \|_{p}$. Thus, minimizing the mutual loss has the same goal as minimizing the quantization loss with lower significance. For simplicity, we will only consider the quantization and the adversarial losses in the following discussion and experiment.

\vspace{-3mm}
\subsection{The Lipschitz Constant and Matrix Norm}

To derive the bound of the aforementioned errors, we introduce the matrix norm as a metric:
\vspace{-2mm}
\begin{equation}
    \label{eq:6_Both}
    \left \| A \right \|_{p}=\sup_{x: \left \| x \right \|_p = 1} \left \| Ax \right \|_p=\sup_{x \neq 0} \frac{\left \| Ax \right \|_p}{\left \| x \right \|_p}.
    \vspace{-1mm}
\end{equation}

Please note that Eq.~(\ref{eq:6_Both}) has the same form as Eq.~(\ref{eq:2_Q}), which means the matrix norm of any matrix $A$ is also the Lipschitz constant of $A$.
We use the Lipschitz constant (the matrix norm) as a metric for quantization optimization based on the following two reasons: (1) According to the definition, the Lipschitz constant is the upper bound of the matrix-vector multiplication. This can be used to upper-bound the errors and give us theoretical guarantee of our method's effectiveness. (2) Compared to testing all types of adversarial examples to fully evaluate models' robustness, using the Lipschitz constant consumes less computation and time and is also input-independent. Faster evaluation could let us find the optimal quantization parameters with affordable overhead.

% \vspace{-3mm}
% \subsection{Error Tolerance}

If we change the quantization bitwidth or switch the quantization method in layer $l$, different quantization settings will result in different $\Delta W^l$. Let us assume the two quantization settings are $q_1$ and $q_2$, respectively. Then the quantization error difference between $q_1$ and $q_2$ in the output of layer $l$ is: $(\Delta W_{q_1}^l-\Delta W_{q_2}^l)x^l$. 
We may further define $L_1=\left \| \Delta W_{q_1}^l \right \|_p$, $L_2=\left \| \Delta W_{q_2}^l \right \|_p$, $\Delta L=\left \| \Delta W_{q_1}^l-\Delta W_{q_2}^l \right \|_p$. According to the triangle inequality: 
\vspace{-1mm}
\begin{equation}
    \label{eq:7_Both}
    \Delta L \geq \left | L_1-L_2 \right |.
    \vspace{-1mm}
\end{equation}

Eq.~(\ref{eq:7_Both}) indicates that instead of computing $\Delta L$, we can use the difference between the Lipschitz constants of the quantization weight errors to estimate the quantization error tolerance difference when we change any quantization parameters.

%% file: 05_method.tex
% \vspace{-4mm}
\section{Methodology}
% \vspace{-2mm}

In this work, we will minimize both adversarial and quantization loss simultaneously.
We give the pseudo code of the layer-wise adversarial-aware quantization method in Algorithm~\ref{LQ_Alg}. After we adversarially train a model, we attempt to quantize the model layer by layer, starting from an extremely low quantization bit first (e.g., binary quantization). We continue increasing quantization bit-width until both the following criteria are met: (1) The Lipschitz constant of the quantized weight $W_{q}[i]$ is below $threshold_{q, i}$, and (2) The Lipschitz constant of the quantization error in weight $W[i]-W_{q}[i]$ is below $threshold_{\Delta, i}$. We try both linear and logarithmic quantization in each layer, and record the best quantization bit and the corresponding quantized weight in both settings, respectively. In Step 3, we compare the best results for each layer and decide which quantization method to be used and the quantization bit-width can also be determined. In the final step, we test the robustness of our quantized model against various adversarial attacks.

In summary, our layer-wise adversarial-aware quantization method introduces acceptable overhead, and it is highly extensible, as additional quantization options (such as new quantization bits and new quantization methods) can be easily added to the algorithm without changing its structure. By using this algorithm, we can theoretically determine an optimal level of our quantization method before testing and save time.

\setlength{\textfloatsep}{2pt}
\begin{algorithm}[bt]
\small
\DontPrintSemicolon
\SetAlgoLined
\SetNoFillComment
\LinesNotNumbered
\footnotesize
% \small
% \hline
% \rule{\textwidth}{0.8pt}
\caption{The Layer-wise Adversarial-aware Quantization Algorithm.}
% \vspace{-2mm}
% \rule{\textwidth}{0.8pt}
% \hline
\label{LQ_Alg}
\SetAlgoNoEnd
\textbf{1. Adversarial Training:} $model.adv\_train(samples, labels)$ 

\textbf{2. Metrics Evaluation:}  \tcp*{L: The model's total layer number.}

% $L=model.total\_layers()$

\For {$i$ in $L$}{ 
    % \tcc{0: Linear quant., 1: Log quant.}
    
    $q[0][i]=q[1][i]=1;$ \tcp*{0: Linear quant., 1: Log quant.}
    
    $Method=[``linear", ``log"]; W[i]=model.layer(i);$
    
    \For {$j$ in 2}{
        $W_q[i]=quant(W[i], quant\_method=Method[j], quant\_bit=q[j][i])$
    
        \While{$Norm(W_q[i])>threshold_{q, i} || Norm(W[i]-W_q[i])>threshold_{\Delta, i}$}{
            $q[j][i]++; W_q[i]=quant(W[i], quant\_method=Method[j], quant\_bit=q[j][i]);$
        }
    
        $W_{q, j}[i]=W_q[i];$ \tcp*{Record the result.}
    }
}

\textbf{3. Quantization Setting Selection:}

\For {$i$ in $L$}{
    \If{$Norm.eval(W[i], W_{q, 0}[i])<Norm.eval(W[i], W_{q, 1}[i])$} {$model.layer(i)=W_{q, 0}[i]$ \tcp*{Choose linear quant.}
    } \Else{$model.layer(i)=W_{q, 1}[i]$ \tcp*{Choose log quant.}
    }
}

\textbf{4. Test the Quantized Model:} $model.test(test\_samples, test\_labels)$

% $count \leftarrow 0;$

% \For {$k$ in $selected\_index$}{
% $examples[count++] \leftarrow generate(samples[k],labels[k], $\\$margin[k],adjacent[k],random\_dirs);$}

% $[retrain\_examples,retrain\_labels] \leftarrow shuffle([samples, labels],$\\$[examples,labels[selected\_index]]);$

% \textbf{5. Retraining and Testing:}\\
% \tcc{Retrain on the trained model.}
% $model.train(retrain\_examples, retrain\_labels)$ 

% $model.test(test\_samples, test\_labels)$ \tcp*{Test.}
\end{algorithm}

% \vspace{-4mm}
\subsection{Threshold and Evaluation Criteria Selection}
% \vspace{-2mm}

In Algorithm~\ref{LQ_Alg}, the two thresholds $threshold_{q, i}$ and $threshold_{\Delta, i}$ and the evaluation criterion $Norm.eval(\cdot)$ are manually selected before running the algorithm. For the threshold selection, based on the definition of the Lipschitz constant and the error amplification effect~\cite{liao2018defense}, we know that if the Lipschitz constant of $W^i$ or $\Delta W^i$ is greater than 1, the output error of layer $i$ may be greater than its input error. On the other hand, restraining all the Lipschitz constants of $W^i$ or $\Delta W^i$ to be less than or equal to 1 may be impossible or will result in high quantization bit-width. During our experiments, we find that:
\vspace{-2mm}
\begin{subequations}
\begin{align} 
    threshold_{q, i} &= 110\% \times L_{W^i}, \\
    threshold_{\Delta, i} &= \left\{\begin{matrix}
    0.3, & if \; L_{W^i}<3.0\\ 
    1.5, & if \; L_{W^i}\geq 3.0
    \end{matrix}\right., (i = 1, 2, ..., L),
    \vspace{-2mm}
\end{align}
\end{subequations}
gives good results. For the evaluation criteria, the algorithm will choose the quantization setting which has lower $L_{\Delta W^i}$. Please note that these criteria are selected empirically and we will not prove the rationality of how we choose these criteria in this work.

%% file: 06_experiment.tex
\vspace{3mm}
\section{Experimental Results}
% \vspace{-2mm}
\label{experiment}

We implement our method with two neural network structures on two datasets: a four-layer CNN on MNIST and a wide ResNet-32 on CIFAR-10. 
% We mainly focus on the performance of adversarially-trained models. 
We conduct both gray- and white-box attacks, with larger gray-box iterative adversarial strength than white-box iterative adversarial strength, as white-box attacks are stronger than gray-box attacks.

% \vspace{-4mm}
\subsection{Robustness Evaluation}

We show the accuracy under various gray- and white-box attacks on adversarial MNIST and CIFAR-10 models in Table~\ref{tab:MNIST_both} and Table~\ref{tab:CIFAR_both}, respectively\footnote{We consider Carlini \& Wagner (CW-L2)~\cite{carlini2017towards}, fast gradient sign method (FGSM)~\cite{goodfellow2014explaining}, projected gradient descent (PGD)~\cite{madry2017towards}, basic iterative method (BIM)~\cite{Kurakin2017ICLR}, momentum iterative method (MIM)~\cite{dong2017boosting}, Gaussian noise, and random noise (with a uniform distribution). The Adv. model is the same as in Figure~\ref{fig:motivation}.}. Here, FGSM (w/s) represents FGSM with weak or strong strengths, and the accuracy pairs are in the form of ``gray-/white-box accuracy''. For different models, ``$Q_1$'' is 3-bit linear quantization, ``$Q_2$'' is 3-bit logarithmic quantization, and ``LQ'' uses the layer-wise adversarial-aware quantization method. 

We check the final quantization settings decided by the layer-wise adversarial-aware quantization method. For MNIST dataset, the LQ model is different with the $Q_2$ model only in the first layer (linear quantization vs. logarithmic quantization) and the third layer (4-bit vs. 3-bit). For CIFAR-10 dataset, all layers of the LQ model use logarithmic quantization, while its layers $7^{th}-11^{th}, 17^{th}$, and $19^{th}-21^{st}$ use 2-bit quantization, layers $1^{st}, 2^{nd}$, and $32^{nd}$ use 4-bit quantization, and the other layers use 3-bit quantization.

\begin{table*}[tb]
\centering
\caption{The accuracy (\%) under gray- and white-box attacks on the Adv. MNIST models.}
\vspace{1mm}
\label{tab:MNIST_both}
\resizebox{\textwidth}{!}{%
\begin{tabular}{lccccccccc}
\toprule
Models      & Clean            & CW-L2            & FGSM (w)         & FGSM (s)         & PGD              & BIM              & MIM              & Gaussian              & Random              \\ \cmidrule(lr){1-1} \cmidrule(lr){2-2} \cmidrule(lr){3-8} \cmidrule(lr){9-10}
Adv.          & 98.4          & 96.5/94.0          & 98.0/98.0          & 96.3/96.2          & 96.7/94.4          & 96.3/93.6          & 96.0/93.5          & 77.2                & 98.4 \\ \cmidrule(lr){1-1} \cmidrule(lr){2-2} \cmidrule(lr){3-8} \cmidrule(lr){9-10}
Adv. ($Q_1$)      & 42.7          & 36.4/42.3          & 43.5/41.3          & 39.9/36.4          & 42.4/40.2          & 42.0/38.4          & 41.3/37.6     & 17.6     & 43.0          \\ 
Adv. ($Q_2$)   & 93.5          & 84.3/92.1          & 92.2/91.3          & 84.8/84.5          & \textbf{86.0}/81.6          & 84.5/80.7          & 83.3/79.0     & 61.9     & 93.1          \\ 
Adv. (LQ)     & \textbf{94.3}          & \textbf{89.6/93.0}          & \textbf{93.3/92.5}          & \textbf{85.9/86.2}           & \textbf{86.0/82.4}           & \textbf{84.9/81.7}           & \textbf{84.0/79.2}     & \textbf{75.4}     & \textbf{94.0}           \\ \bottomrule
\end{tabular}%
}
% \vspace{-2mm}
\end{table*}

\begin{table*}[tb]
\centering
\caption{The accuracy (\%) under gray- and white-box attacks on the Adv. CIFAR-10 models.}
% \vspace{1mm}
\label{tab:CIFAR_both}
\resizebox{\textwidth}{!}{%
\begin{tabular}{lccccccccc}
\toprule
Models      & Clean            & CW-L2            & FGSM (w)         & FGSM (s)         & PGD              & BIM              & MIM              & Gaussian              & Random              \\ \cmidrule(lr){1-1} \cmidrule(lr){2-2} \cmidrule(lr){3-8} \cmidrule(lr){9-10}
Adv.          & 87.3          & 86.3/54.2          & 86.2/74.7          & 68.6/36.8          & 80.6/46.8          & 82.1/45.0          & 76.7/48.6          & 70.3                & 84.9 \\ \cmidrule(lr){1-1} \cmidrule(lr){2-2} \cmidrule(lr){3-8} \cmidrule(lr){9-10}
Adv. ($Q_1$)      & 19.8          & 20.0/19.3          & 19.4/18.7          & 16.2/16.1          & 16.5/17.7          & 17.0/17.6          & 16.7/18.1     & 15.6     & 17.8          \\ 
Adv. ($Q_2$)   & 33.9          & 33.7/31.6          & 33.1/31.3          & 23.9/21.1          & 27.1/28.5          & 30.0/28.4          & 25.1/28.7     & 23.0     & 30.2          \\ 
Adv. (LQ)     & \textbf{81.8}          & \textbf{81.6/60.1}          & \textbf{80.9/69.2}          & \textbf{68.2/34.7}           & \textbf{75.7/48.6}           & \textbf{77.4/48.5}           & \textbf{70.8/50.3}     & \textbf{70.8}     & \textbf{79.4}           \\ \bottomrule
\end{tabular}%
}
\end{table*}

For MNIST results (Table~\ref{tab:MNIST_both}), the LQ model and the $Q_2$ model have similar performance. This is because their quantization settings are similar. For CIFAR-10 results (Table~\ref{tab:CIFAR_both}), we find that the LQ model outperforms the $Q_1$ and $Q_2$ model by a large margin, and even outperforms the full-precision model under white-box iterative attacks. These results show that the layer-wise adversarial-aware quantization method can effectively minimize both the adversarial and the quantization losses. Calculating the Lipschitz constant of quantized weights and quantized weight errors as metrics can successfully guide the selection of quantization parameter settings, which gives better robustness performance. Overall, the LQ models always have the best robustness among all considered models, and they have acceptable accuracy drop compared with the full-precision adversarial models.

% \vspace{-3mm}
\subsection{Further Validation with the Lipschitz Constant}
% \vspace{-2mm}

To further evaluate our method, we also check the Lipschitz constant of quantized weights and quantized weight errors in every layer. According to Table~\ref{tab:Lipschitz_both}, the Lipschitz constant (p=2) of our method has a lower mean value and smaller standard deviation compared with the second-best model $Q_2$. Please note that the results shown in Table~\ref{tab:Lipschitz_both} are calculated based on layers with different quantization settings (which means layers with 3-bit logarithmic quantization for both the LQ and $Q_2$ models are ignored). This is consistent with our theoretical derivation in Section~\ref{theoretical_both} and also proves the effectiveness of our method: a lower Lipschitz constant indicates better robustness, while optimizing layers with large Lipschitz constants improves robustness more.

\begin{table}[tb]
\centering
% \vspace{3mm}
\caption{The means and the standard deviations of the Lipschitz constant of selected layers.}
\label{tab:Lipschitz_both}
\resizebox{0.65\linewidth}{!}{%
\begin{tabular}{lcccc}
\toprule
          & \multicolumn{2}{c}{MNIST}                                   & \multicolumn{2}{c}{CIFAR-10}                                \\ \cmidrule(lr){2-3} \cmidrule(lr){4-5}
          & $L_{W_q}$                      & $L_{\Delta W}$                    & $L_{W_q}$                      & $L_{\Delta W}$                    \\ \cmidrule(lr){1-1} \cmidrule(lr){2-5}
Adv. ($Q_2$) & $6.38 \pm 4.99$ & $1.83 \pm 1.65$ & $1.03 \pm 1.16$ & $0.26 \pm 0.39$ \\
Adv. (LQ) & $6.15 \pm 4.86$ & $0.91 \pm 0.74$ & $0.96 \pm 1.06$ & $0.26 \pm 0.13$ \\ \bottomrule
\end{tabular}%
}
\end{table}

%% file: 07_conclusions.tex
% \vspace{-6mm}
\section{Conclusions}
% \vspace{-4mm}

In this work, we discover that with careful quantization setting selection, the robustness of quantized adversarial models can be regained. Then we theoretically derive the adversarial and the quantization losses, and choose the Lipschitz constant as the error sensitivity metric. We also propose a layer-wise adversarial-aware quantization method to minimize both the adversarial and quantization loss in quantization scenarios. Our experimental results show that the robustness of quantized adversarially-trained models can be restored to the same level as their full-precision counterparts, and the Lipschitz constant also proves the effectiveness of our method. These findings may provide helpful guidance to future research on quantized neural network robustness and neuromorphic computing applications. 